\documentclass{article}

\PassOptionsToPackage{numbers, compress}{natbib}

\usepackage[preprint]{neurips_2024}

\usepackage{enumitem} 
\usepackage{amsmath}
\usepackage{graphicx} 
\usepackage[utf8]{inputenc} 
\usepackage[T1]{fontenc}    
\usepackage{hyperref}       
\usepackage{url}            
\usepackage{booktabs}       
\usepackage{amsfonts}       
\usepackage{nicefrac}       
\usepackage{microtype}      
\usepackage{xcolor}         
\usepackage{subcaption}
\usepackage{caption}
\usepackage{wrapfig} 
\usepackage{graphicx} 
\graphicspath{ {./figures/} }
\usepackage{amsmath}
\usepackage{algorithm}
\usepackage{algpseudocode}
\usepackage{multirow}
\usepackage{amssymb}
\usepackage{colortbl}

\title{USP: A Unified Sequence Parallelism Approach for Long Context Generative AI}

\author{
Jiarui Fang \\
  Tencent \\
  \texttt{jiaruifang@tencent.com} \\
   \And
  Shangchun Zhao \\
  Tencent \\
  \texttt{doctzhao@tencent.com} \\
}

\begin{document}

\maketitle

\begin{abstract}
Sequence parallelism (SP), which divides the sequence dimension of input tensors across multiple computational devices, is becoming key to unlocking the long-context capabilities of generative AI models.
This paper investigates the state-of-the-art SP approaches, i.e. DeepSpeed-Ulysses and Ring-Attention, and proposes a unified SP approach, which is more robust to transformer model architectures and network hardware topology.
This paper compares the communication and memory cost of SP and existing parallelism, 
including data/tensor/zero/pipeline parallelism, 
and discusses the best practices for designing hybrid 4D parallelism involving SP.
We achieved 47\% MFU on two 8xA800 nodes using SP for the LLAMA3-8B model training using sequence length 208K.
Our code is publicly available at \url{https://github.com/feifeibear/long-context-attention}.
\end{abstract}


\section{Introduction}
The field of artificial intelligence is witnessing a trend as the context length in generative AI models grows ever longer. 
Claude has pioneered this trend in large language models (LLMs) by extending the sequence length to 100K tokens. 
Following closely in its footsteps, OpenAI's GPT-4 has expanded the context length to 128K tokens.
The advent of multi-modality models is propelling this trend forward, with Gemini 1.5 Pro boasting a context length of a staggering 10 million tokens, and OpenAI's Sora, a Diffusion Model, accommodating at least 1 million visual tokens.
These breakthroughs underscore the imperative for generative AI techniques to adeptly handle a larger context length.

Sequence Parallelism (SP), a technique that partitions input sequences, has emerged as a promising approach for the training or inference of longer sequences. 
Following an initial exploration period of two years, by the latter of 2023, two landmark works, DeepSpeed-Ulysses~\cite{jacobs2023deepspeed} and Ring-Attention~\cite{liu2023ring}, marked the maturation of the SP technique. 
DeepSpeed-Ulysses maintains constant communication volume when sequence length and compute devices are increased proportionally, 
while Ring-Attention hides P2P communication costs introduced by SP through overlapping computation and communication. 
However, challenges remain, such as the SP parallel degree of DeepSpeed-Ulysses is limited to less than the number of attention heads, and the computational efficiency of Ring-Attention degrading due to the subdivision of matrix multiplications.
These limitations currently hinder the broader adoption of Sequence Parallelism in distributed Transformer computation.

In this paper, we delve deeper into the realm of SP. 
We begin by highlighting that Ulysses and Ring are not mutually exclusive approaches; they can be combined through a hybrid parallel strategy to mitigate their drawbacks.
Then, we discussed the relationship between SP and data/tensor/zero/expert/pipeline parallelism. 
The most complex among these is the relationship between SP and tensor parallelism. Since tensor parallelism also has its specific sequence parallel optimizations to reduce activation memory cost~\cite{korthikanti2023reducing}.
For each parallelism approach, whether SP should replace it or is there some issue to using SP with it together, remains an open question.
After addressing these questions, we have provided a list of best practices for building a 4D hybrid parallelism system.

The primary contributions of this paper include:

\begin{itemize}
    \item We propose a unified sequence parallel method that integrates DeepSpeed-Ulysses and Ring-Attention, overcoming the shortcomings of both and demonstrating greater robustness to model architecture and network hardware.
    \item We systematically analyze the application of SP in conjunction with Tensor Parallelism, ZeRO, and Pipeline Parallelism as 4D parallelism, and provide a list of best practices to apply SP.
\end{itemize}

\section{Sequence Parallelism Approaches}
Before delving into the concept of SP, let's review the computational process of the standard Transformer Block. 
The Notion used in this paper is shown in Table~\ref{tab:transformer_params}, where $d = hc\times hs$.

\begin{table*}[!h]
\centering
\small
\begin{tabular}{cccccc}
\toprule
$L$ & Sequence Length & $d$ & Hidden Dimension & $hc$ & Head Count \\
\hline
$hs$ & Head Size & $bs$ & Batch Size & $N$ & Device Number \\
\hline
\end{tabular}
\caption{Notation for Transformer Parameters}
\label{tab:transformer_params}
\end{table*}

Given input sequences $Q, K, V \in \mathbb{R}^{L \times d}$, where $L$ is the sequence length and $d$ is the head dimension, we compute the matrix of outputs as follows:

\begin{equation}
\text{Attention}(Q, K, V) = \text{softmax}\left(\frac{QK^T}{\sqrt{d}}\right)V,
\end{equation}

Each self-attention sub-layer is accompanied by a feedforward network (FFN), which is applied to each position separately and identically.

\begin{equation}
\text{FFN}(x) = \max(0, xW_1 + b_1)W_2 + b_2.
\end{equation}

Compared to Tensor Parallelism (TP)~\cite{shoeybi2019megatron} and ZeRO~\cite{ren2021zero}, the research on SP for Transformer models has been relatively underdeveloped for a long time. 
The challenge lies in the characteristic of attention computation, where the sequence dimension serves as a common dimension in the matrix multiplication after softmax, making it difficult to partition the tensors and distribute the computation across multiple nodes after slicing the sequence dimension.

\begin{figure}[htbp]
\centering
\includegraphics[width=0.7\textwidth]{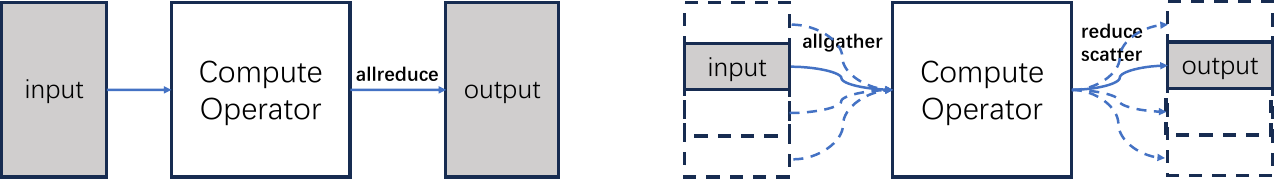}
\caption{The Principle of Megatron-LM Sequence Parallelism.}
\label{fig:allreduce_opt}
\end{figure}

Early attempts~\cite{li2021sequence, li2023colossal, li2023lightseq} at SP were not successful, often leading to redundant memory consumption~\cite{li2021sequence} and inefficient communication pattern~\cite{li2023lightseq}. 
For long input sequences, the best practice is to adopt the \textit{Sequence Parallelism} of Megatron-LM. 
This method optimizes the AllReduce operation of TP, reducing the memory cost of activations while maintaining the same communication overhead. 
As shown in Figure~\ref{fig:allreduce_opt}, the principle of Megatron-LM \textit{Sequence Parallelism} is the similar to ZeRO-2~\cite{ren2021zero}.
It replaces the AllReduce operation on replicated tensors (the left figure) into equivalent allgather and reduce-scatter operations on partitioned data (the right figure). 
Since an AllReduce operation is exactly a combination of Allgather and ReduceScatter, the communication cost remains the same.
The size of the input and output tensors is reduced by a factor of 1/$N$ across $N$ computational devices. 
Because of the sequence dimension in input/output tensors is partitioned, it is named as \textit{Sequence Parallelism}. 
However, this form of \textit{Sequence Parallelism} cannot be used independently without tensor parallelism and communication volume remains constant regardless of the degree of parallelism.

The maturity of the standalone SP technology is marked by the publication of two milestone papers in late 2023.
The DeepSpeed-Ulysses~\cite{jacobs2023deepspeed}, named as \textbf{SP-Ulysses}, and Ring-Attention~\cite{liu2023ring}, named as \textbf{SP-Ring}, solving the longstanding memory and communication issues inherent to SP from two distinct perspectives.
For both methodologies, each computational device is allocated a distinct segment of the $Q$ (query), $K$ (key), $V$ (value), and $O$ (output) tensors, which are segregated along the dimensions of the sequence. There is no redundancy in their storage between devices, which is a primary distinction from the early SP design~\cite{li2021sequence}.

\begin{figure}[htbp]
\centering
\includegraphics[width=0.8\textwidth]{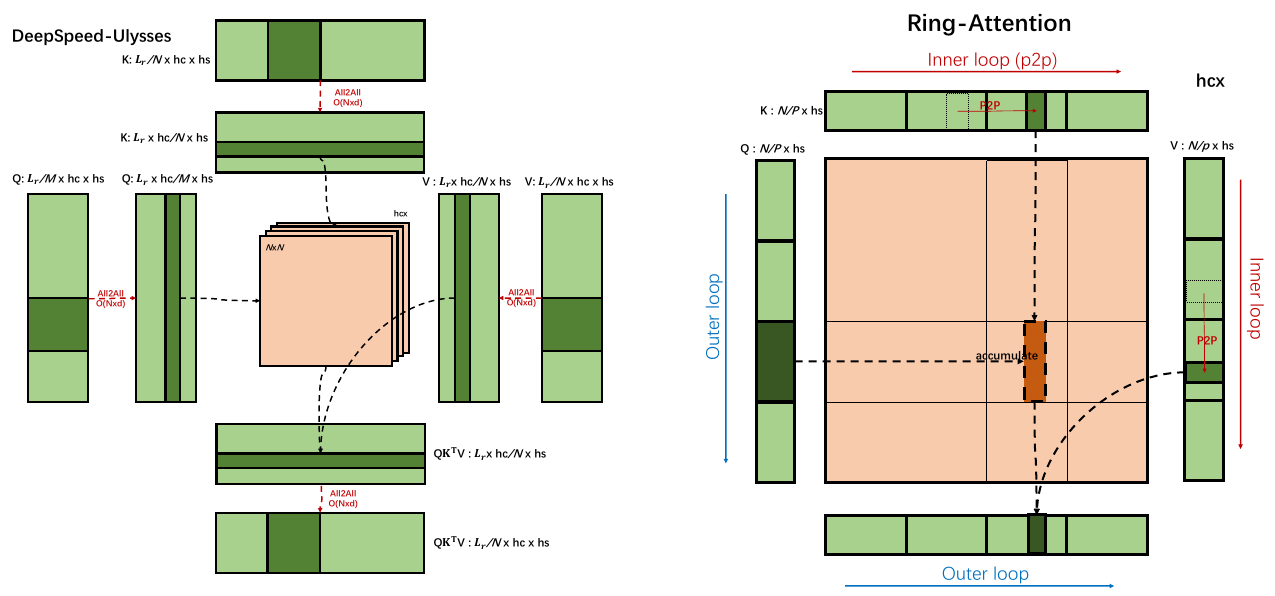}
\caption{SP-Ulysses and SP-Ring.}
\label{fig:sp_ulysses_ring}
\end{figure}

SP-Ring can be viewed as a distributed version of FlashAttention~\cite{dao2022flashattention}. 
As shown in the right part of Figure~\ref{fig:sp_ulysses_ring}, SP-Ring employs a nested two-level loop that orchestrates the communication and computation in a blockwise fashion.
When computing for the blocks of the tensor $O$ segment, if the required tensor $K$ and $V$ blocks are not locally available, Peer-to-Peer (P2P) communication is utilized to fetch them from other devices. 
Communication can be organized in a Ring fashion, where each device simultaneously sends and receives $K, V$ blocks, allowing communication to overlap computation.

SP-Ulysses leverages All2All communication for segments of the $Q$, $K$, $V$, and $O$ tensors, as shown in the left part of Figure~\ref{fig:sp_ulysses_ring}. 
After the All2All operation, the partitioning of these four tensors changes from the sequence dimension $L$ to the dimension of the attention number heads $hc$. 
Therefore, the computation of $softmax(QK^T)V$ for each attention head is maintained in its entirety, and can be implemented using the under the hood Attention operator library, like FlashAttention.

\section{Unified Ulysses-Ring Sequence Parallelism}
\label{sec:sp-unified}
Currently, both SP-Ulysses and SP-Ring are facing certain issues that hinder their effectiveness in practical applications.

SP-Ulysses is sensitive to the number of attention heads. 
The parallelism degree~\footnote{The parallelism degree is the number of devices that participate in parallel computation. In other words, it is the number of processes in a parallel process group.} of DS-Ulysses cannot exceed the number of attention heads $hc$. 
Consequently, it is not suitable for the GQA (Grouped Query Attention)~\cite{ainslie2023gqa} and MQA (Multi-Query Attention)~\cite{shazeer2019fast} scenarios. 
For instance, Llama3-8B employs GQA with a KV head number of 8, which means that when using DS-Ulysses SP, the maximum SP degree is 8. 
However, if MQA is used and the KV head number is 1, DS-Ulysses will not function.
In addition, since Tensor Parallelism also requires division across the $hc$ dimension, SP-Ulysses, and TP are in conflict.

SP-Ring is inefficient in computation and communication. 
Ring-Attention segments the $Q, K, V, O$ tensors into smaller blocks, which can lead to a decrease in computation efficiency of the fused operator $Softmax(QK^T)V$. 
Even if communication and computation fully overlap, 
the total execution time lags behind that of DS-Ulysses. 
When using a causal mask has not been addressed, the DS-Ring has a load-unbalancing issue.
DS-Ring does not impose any restrictions on $hc$.

SP-Ulysses and SP-Ring are currently considered alternative strategies for SP, with the choice of one precluding the other. 
This point of view was reinforced by the SP-Ring authors in their ICLR open review rebuttal\footnote{https://openreview.net/forum?id=WsRHpHH4s0\&noteId=HIY0tae4Gz}. 
Currently, Megatron-DeepSpeed utilizes SP-Ulysses, while Megatron-LM opts for SP-Ring for its SP implementation. 
However, we claim that, rather than viewing them as rivals, they can work together as a unified SP approach.

As shown in Algorithm~\ref{alg:unified_sp}, SP-Ring and SP-Ulysses are organized in a hybrid parallel manner named as \textbf{USP-Attention} to work together in partitioning the sequence dimension.
The SP process group is segmented into two orthogonal process group sets: a set of SP-Ring process groups and a set of SP-Ulysses process groups.
For a more intuitive understanding, an SP process group can be viewed as a 2D mesh, SP-Ring operates across each column of the mesh, while SP-Ulysses runs across each row.
For example, a process group containing 8 processes could viewed as $2\times4$, where a SP-Ulysses process group $ulysses\_pg$ of size 2 and a SP-Ring process group $ulysses\_pg$ of size 4.
This is the same as how data parallelism and tensor parallelism process groups are partitioned.

The inputs to the \textbf{USP-Attention} include the segment of the $Q, K, and V$ tensors after being partitioned along the sequence dimension and the output is a tensor $O$ shard. 
The size of a tensor segment is ($bs$, $L/N$, $hs$, $hd$). 
Note that when using MAQ, the shape of the $hc$ for the $K$ and $V$ tensor segments differs from that of the Q tensor.
During forward propagation, $scatter\_idx$ is set to 1, and $gather\_idx$ is set to 2. 
 AllToAll4D merges the dimension $L$ and partitions the dimension $hc$ of the tensors $Q$, $K$, and $V$, transforming them into ($hc/N$, $bs$, $L$, $d$). They also partition the $O$ tensor along the $L$ dimension and merge the $hc$ dimension.
During backward propagation, $scatter\_idx$ is set to 2, and $gather\_idx$ is set to 1.

\begin{algorithm}
\caption{Unified Sequence Parallelism Attention Implementation}
\label{alg:unified_sp}
\begin{algorithmic}[1]
    \Function{\textbf{USP-Attn}}{$ulysses\_pg$, $ring\_pg$, $Q$, $K$, $V$, $scatter\_idx$, $gather\_idx$}
        \State $Q \gets \textbf{AllToAll4D}{(Q, scatter\_idx, gather\_idx, group=ulysses\_pg)}$
        \State $K \gets \textbf{AllToAll4D}{(K, scatter\_idx, gather\_idx, group=ulysses\_pg)}$
        \State $V \gets \textbf{AllToAll4D}{(V, scatter\_idx, gather\_idx, group=ulysses\_pg)}$
        \State $O \gets \textbf{LoadBalance-RingAttention}(Q, K, V, group=ring\_pg)$
        \State $O \gets \textbf{AllToAll4D}{(O, gather\_idx, scatter\_idx, group=ulysses\_pg)}$
        \State \Return $O$
    \EndFunction
\end{algorithmic}
\end{algorithm}

\begin{wrapfigure}{r}{0.6\textwidth} 
\centering
\includegraphics[width=0.59\textwidth]{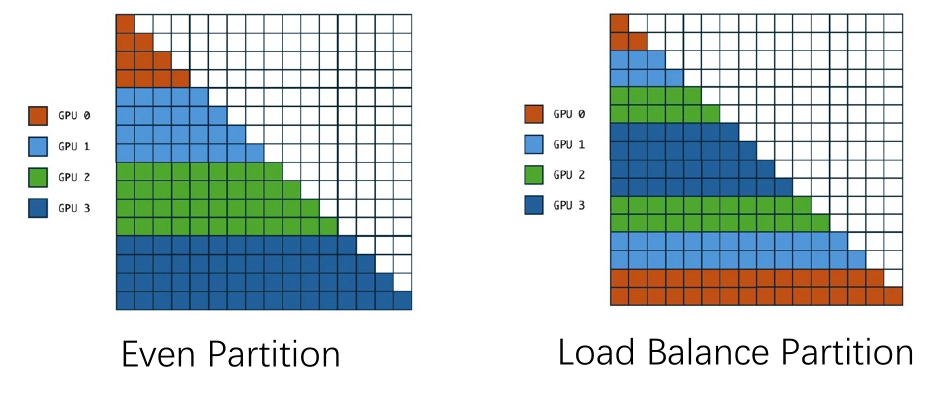}
\caption{Loading Balancing for SP-Ring.}
\label{fig:lb}
\end{wrapfigure}

The vanilla SP-Ring introduces load-unbalancing issues when applying causal attention, as only the lower triangular matrix of $QK^T$ needs to be computed. 
As shown in Figure~\ref{fig:lb}, if the sequence dimension is divided evenly, the computational tasks are not evenly distributed among the devices. 
As shown in the left side of the figure, on 4 GPUs the computation load of GPU3 is nearly 7 times that of GPU0.

The solution to the load-unbalancing issue is to reorder the input sequence tokens along the sequence dimension, as depicted on the right side of Figure~\ref{fig:lb}. 
In the figure, 
the input sequence consists of 16 tokens. 
Under even partition, GPU0 processes tokens 0-3, while GPU3 handles tokens 12-15. After the reorder operation for load balance partitioning, GPU0 now processes tokens 0, 1, 14, 15, and GPU3 processes tokens 5,6,11,12.  
The workloads handled by each GPU are perfectly balanced, which is a superior solution to the striped attention~\cite{brandon2023striped}.

The aforementioned load balancing method is quite straightforward and a similar implementation has already been applied in Megatron-LM. 
We present its application for Unified SP method.
The Algorithm~\ref{alg:extract} extracts and reorders the input sequence from the global input sequence in hybrid Ulysses and Ring parallelism.
For the positional encoding only involves element-wise operations, i.e. RoPE~\cite{su2024roformer}, 
the same reordering operation also needs to be applied to the positional encoding parameters of the model. 
Since the operation is applied to the model's input token sequence, which is an integer vector of length $bs*L$, the additional overhead for load balancing is negligible.

\begin{algorithm}[htbp]
\caption{Prepare Load Balance Sequence Segment from The Global Input Sequence}
\label{alg:extract}
\begin{algorithmic}[1]
\Function{LocalBalanceLocalSeq}{$seq$, $ring\_process\_group$, $ulysses\_process\_group$}
    \State $ring\_degree \gets ring\_process\_group.get\_world\_size()$
    \State $ring\_rank \gets ring\_process\_group.get\_rank()$
    \State $ulysses\_rank \gets ulysses\_process\_group.get\_rank()$
    \State $seq\_chunks \gets seq.chunk(2 \times ring\_degree)$
    \State $reorder\_seq \gets concat([seq\_chunks[r\_rank], seq\_chunks[2 \times rd - r\_rank - 1]])$
    \State $local\_seq \gets reorder\_seq.chunk(ud)[u\_rank]$
    \State \Return $local\_seq$
\EndFunction
\end{algorithmic}
\end{algorithm}

Unified SP is highly flexible and robust, allowing for the various combinations of the Ulysses degree and the Ring degree, as long as the product of the Ulysses degree and the Ring degree equals the SP degree.
When the Ulysses degree equals $N$, it becomes SP-Ulysses, and when the ring degree equals $N$, it becomes SP-Ring.
Therefore, it covers the ability of both SP-Ulysses and SP-Ring.

\begin{figure}[htbp]
\centering
\includegraphics[width=0.8\textwidth]{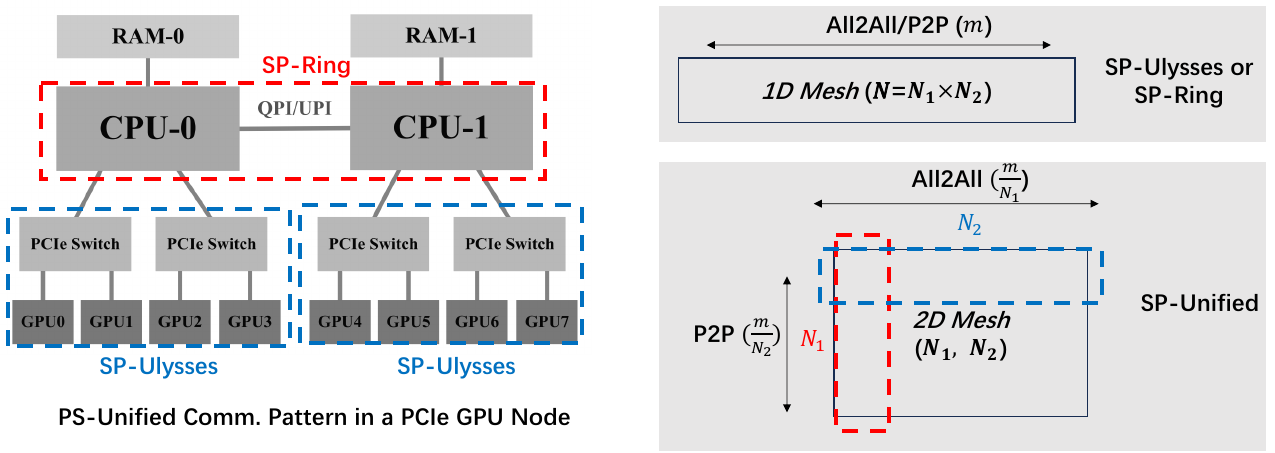}
\caption{The Unified-SP is more robust to network hardware topology.}
\label{fig:spnet}
\end{figure}

Firstly, Unified SP can remove the head number limitation of SP-Ulysses. For example, to run llama3-8B~\footnote{https://github.com/meta-llama/llama3} with $hc$=8 in a 16-degree SP, one can set the Ulysses degree to 8 and the ring degree to 2. 
Secondly, Unified SP allows lower bandwidth and topology requirements for network infrastructure by offering a more robust communication pattern. 
By setting the Ulysses degree to a value between 1 and N, the Attention communication pattern will be a mix of P2P and All2All as shown in Figure~\ref{fig:spnet}.
Such a communication pattern is particularly well-suited for heterogeneous communication networks, allowing All2All operations to operate in high-bandwidth interconnections while asynchronous P2P communications operate in lower-bandwidth sections. 
This is applicable to scenarios such as a node in which GPUs are connected via PCIe Switch or a cluster of GPU nodes between which the network is Ethernet connected.

\textbf{\textit{Tip 1: We suggest using Unified-SP in place of SP-Ring and SP-Ulysses, as it encompasses the capabilities of both while offering additional benefits.}}

\section{SP in 4D Parallelism}
SP, as a newly emerged parallelism method, how to integrate it into the existing hybrid parallelism framework of Data Parallelism (DP), Tensor Parallelism (TP), and Pipeline Parallelism (PP) has not been thoroughly studied. 
This section will analyze the relationships between SP and DP, TP, and PP, and discuss the best practices for designing 4D parallelism involving sequence dimension parallelism.

As shown in Table~\ref{tab:memcommcomp}, we analyze the communication and memory cost of a standard transformer block for different parallelism.
The impact of GQA is not reflected in the table, but we will analyze it later.
Entries in the Communications Params columns indicate the collective communication operations are conducted on the parameters and gradients of the transformer block. 
It includes the parameters/gradients of the weight and bias tensors of 4 Linear Layers in the Self-Attention layer, as well as 2 Linear Layers of the FFN 
 layer, amounting to a volume of $12 \times O(d^2)$ elements in the GPT-2 model.
Note that llama3 and llama2 has $9.37 \times O(d^2)$ elements since the intermediate size is lower.
Entries in Communications Act(ivatiom) columns indicate the collective communications are conducted on a single hidden states tensor belonging to activations, containing $bs \times L \times d$ elements.
The memory cost is broken down into model parameters/gradients (\textbf{P/G}), Optimizer States (\textbf{OS}), and intermediate activation tensors (\textbf{Act}).

The Cost of the Communication represents the bandwidth requirements. 
It is calculated by the product of the communicated element number by an \textit{algorithm bandwidth (algobw)} factor related to the algorithm of collective communication~\footnote{https://github.com/NVIDIA/nccl-tests/blob/master/doc/PERFORMANCE.md}.
For the collective communication algorithm of AllReduce, AllGather, ReduceScatter, and AllToAll, the respective algobw factors are \( 2\frac{n-1}{n} \), \( \frac{n-1}{n} \), \( \frac{n-1}{n} \), and 1. 
In the table, we approximate the term $O(\frac{n-1}{n})$ to $O(1)$ for simplicity.

The table is built for the mix-precision training using the fp16(bf16) format.
The memory requirement for the model parameters and gradients are $P$ and $G$ Bytes.
The Optimizer States (OS), which includes the parameter in fp32, momentum, and variance in the Adam optimizer, is 6 times that of the fp16 parameters.
The memory requirement for the peak activation is $A$ bytes.
The parallel degree is $N$.

\begin{table}[h]
\centering
\scriptsize
\caption{Comparison of Communications and Memory Cost of SP, DP, TP, and ZeRO for a standard transformer block}
\label{tab:memcommcomp}
\begin{tabular}{@{}c|cc|cc|c|cccc@{}}
\toprule
& \multicolumn{4}{c|}{\textbf{Communication (FWD+BWD)}} & {\textbf{Split}} & \multicolumn{3}{c}{\textbf{Memory}} \\
& \textbf{Param} & \textbf{Cost} & \textbf{Act} & \textbf{Cost} & \textbf{Dim} & \textbf{P/G} & \textbf{OS} & \textbf{Act} \\
\midrule
SP-Ulysses & allreduce & $12O(d^2)$ & 8*all2all & $\frac{8}{N}O(bs*L*d)$ & $hc$/$L$ & $P+G$ & $6P$ & $A/N$ \\
\midrule
SP-Ring & allreduce & $12O(d^2)$ & P2P & $4O(bs*L*d)$ & $L$/$L$ & $P+G$ & $6P$ & $A/N$ \\
\midrule
DP & allreduce & $12O(d^2)$ & 0 & 0 & $bs$/$bs$ & $P+G$ & $6P$ & $A/N$ \\
\midrule
\multirow{2}{*}{ZeRO1} &  allgather+ & \multirow{2}{*}{$12O(d^2)$}  & \multirow{2}{*}{0} & \multirow{2}{*}{0} & \multirow{2}{*}{$hc$/$L$} & \multirow{2}{*}{$P+G$} & \multirow{2}{*}{$6P/N$} & \multirow{2}{*}{$A/N$} \\
 &   reducescatter &  & & & & & & \\
\midrule
SP-Unified+ &  allgather+ & \multirow{2}{*}{$12O(d^2)$}  & \multirow{2}{*}{P2P+8*all2all} & \multirow{2}{*}{$\leq4O(bs*L*d)$} & \multirow{2}{*}{$hc$/$L$} & \multirow{2}{*}{$P+G$} & \multirow{2}{*}{$6P/N$} & \multirow{2}{*}{$A/N$} \\
ZeRO1 &   reducescatter &  & & & & & & \\
\midrule
SP-Unified+ &  allgather+ & \multirow{2}{*}{$12O(d^2)$}  & \multirow{2}{*}{P2P+8*all2all} & \multirow{2}{*}{$\leq4O(bs*L*d)$} & \multirow{2}{*}{$hc$/$L$} & \multirow{2}{*}{$P+\frac{G}{N}$} & \multirow{2}{*}{$6P/N$} & \multirow{2}{*}{$A/N$} \\
ZeRO2 &   reducescatter &  & & & & & & \\
\midrule
SP-Unified+ &  2*allgather+ & \multirow{2}{*}{$18O(d^2)$}  & \multirow{2}{*}{P2P+8*all2all} & \multirow{2}{*}{$\leq4O(bs*L*d)$} & \multirow{2}{*}{$hc$/$L$} & \multirow{2}{*}{$\frac{P+G}{N}$} & \multirow{2}{*}{$6P/N$} & \multirow{2}{*}{$A/N$} \\
ZeRO3 &   reducescatter &  & & & & & & \\
\midrule
TP & 0 & 0 & 4*allreduce & $8O(bs*L*d)$ & $hc$/$d$ & $\frac{P+G}{N}$ & $6P/N$ & $\alpha A$\\
\bottomrule
\multirow{2}{*}{TP-sp} & \multirow{2}{*}{0} & \multirow{2}{*}{0} & 6*allgather+ & \multirow{2}{*}{$10O(bs*L*d)$} & \multirow{2}{*}{$hc$/$d$} & \multirow{2}{*}{$\frac{P+G}{N}$} & \multirow{2}{*}{$6P/N$} & \multirow{2}{*}{$A/N$} \\
 &  &  & 4*reducescatter & & & & \\
\bottomrule
\end{tabular}
\end{table}

\paragraph{Data Parallelism (DP):}
In terms of communication cost, SP is inferior to DP. 
Both SP and DP require the allreduce operation on gradients during backward propagation. 
The difference in their communication performance lies in the attention module, where SP introduces additional communications overhead for activations.
When the Ulysses degree is set to be greater than 1, the communication overhead of SP will be larger than that of DP due to the all2all operations. 
When using the Ring method, although the additional P2P communication for attention is overlapped, it introduces extra performance issues. The ideal performance is only to reach the performance of DP without communication for attention. In terms of memory performance, both SP and DP are equivalent, as they all can reduce the activation footprint to 1/$N$.

\textbf{\textit{Tip 2: We suggest prioritizing the use of DP over SP if possible. 
Only when the batch size (bs) is insufficient for partitioning should one consider whether to employ SP.}}

\paragraph{ZeRO:}
ZeRO~\cite{rajbhandari2020zero} is a distributed parameter management method that reduces the storage space requirements of each computing device by sharding the Optimizer States (ZeRO-1), Gradients (ZeRO-2), and Parameters (ZeRO-3) across multiple devices. 
The memory cost for Optimizer States, Gradients, and Parameters is reduced to 1/$N$ of the original. ZeRO can also operate within an SP process group because partitioning along the batch dimension ($bs$) or the sequence dimension ($L$) is equivalent to ZeRO's approach.
ZeRO is working on the unified process group of size $N_{sp}\times N_{dp}$, which combines the SP and DP process groups.

\textbf{\textit{Tip 3: We suggest that when utilizing SP, it should always be used in conjunction wit ZeRO-1/2.}}
One can also consider employing ZeRO-3, and Offload techniques~\cite{ren2021zero, fang2022parallel} to trade off communication cost for memory savings.

\paragraph{Tensor Parallelism (TP):}
The Tensor Parallelism (TP) approach, pioneered by Megatron-LM~\cite{shoeybi2019megatron}, shards the parameters of models across computing devices. 
In the TP part of activation tensors, not all are partitioned and distributed across multiple computational devices. 
Consequently, the memory cost for activations, as reflected in the Table~\ref{tab:memcommcomp}, is denoted by the $\alpha A$, where 0<$\alpha$<1. 
Please refer to the Equation(2) of paper~\cite{korthikanti2023reducing} for a more precise $\alpha$.
TP has been further refined by Megatron-LM Sequence Parallelism~\cite{korthikanti2023reducing}, which replaces the an allreduce in TP
with an allgather and a reducescatter, and therefore reduces the activation memory cost to $A/P$ at the cost of redo two allgathers for attention and FFN.
To distinguish Megatron-LM Sequence Parallelism from Ring and Ulysses Sequence Parallelism, we named it TP-sp here.

As shown in Table~\ref{tab:memcommcomp}, in terms of communication cost, TP-sp is higher than SP-Ulysses and SP-Ring.
Firstly, the communication volume of TP-sp is greater than that of SP-Ring, and the latter can be overlapped with computation.
Additionally, the communication volume of TP-sp does not decrease with an increase in parallelism, whereas SP-Ulysses can achieve this.
Therefore, TP-sp is inferior to any form of sequence parallelism in terms of communication.
The SP has a lower communication cost for activations,  but it requires synchronizing gradients and parameters.
However, the parameter communication volume is small compared to the activation communication volume, and it can be overlapped by computation.
GQA/MQA can reduce SP communication costs, while the communication cost of TP-sp remains unchanged. 
Assuming the GQA group number is $G$, the Ulysses and Ring communication cost for $K$ and $V$ is reduced to 1/$G$, and the activation communication cost is reduced to $\frac{4}{N}O(bs*L*d) + \frac{4}{N}O(bs*L*d/G)$ and $\frac4(bs * L * \frac{d}{G})$.

\textbf{\textit{Tip 4: We suggest that SP has an advantage over TP-sp in terms of communication cost on a large scale. GQA can further reduce the communication cost of SP.
}}

In terms of memory cost, TP-sp holds an advantage over SP. 
Even when SP employs ZeRO-1/2 to align the memory cost of OS and activations with that of TP-sp, the memory cost of the parameter remains more substantial. 
However, when SP employs ZeRO3, it achieves a similar memory cost with TP-sp.
The SP-Ulysses+ZeRO3 is the exact strategy to scale sequence length employed by the authors of DS-Ulysses~\cite{jacobs2023deepspeed}.

Based on the above analysis, under the same model and input configurations~\ref{tab:transformer_params}, simply switching parallelism from TP-sp to SP-Unified does not increase the trainable sequence length; on the contrary,  it is very likely to result in a reduction of the sequence length.

However, SP can still extend the sequence length compared to TP-sp in high parallel degree.
When the sequence is very long, as previously analyzed, the proportion of parameter communication volume in the total communication is relatively small. Therefore, the additional communication overhead of an allgather operation introduced by ZeRO has a limited impact.

\textbf{\textit{Tip 5: We suggest switching TP-sp to SP cannot increase the sequence length in training.
SP+ZeRO3 can train a similar sequence length as TP-sp.
}}

Due to the inherent limitation of TP-sp's parallelism, which is capped at hc, it is not possible to further reduce the memory cost of activations by increasing the TP-sp paralleled degree. 
In contrast, SP can continue to expand its parallel degree by leveraging the SP-Ring technique, thereby enabling the training of larger models on a larger scale.

\textbf{\textit{Tip 6: We suggest a higher degree of SP parallelism, which may need to set a large ring degree when the head number is limited, to train a long sequence across a greater number of computational devices. 
This is an advantage that cannot be achieved with TP-sp approaches.
}}

\paragraph{Pipeline Parallelism (PP):}
PP partitions transformer blocks across layers, so it is complementary with the SP, which partitions tensors inside a transformer block. Therefore, SP and PP are fully compatible. Since SP can form a unified parallel group with ZeRO for collective communication, we believe TP should still be placed at the lowest dimension of the 4D parallel group partitioning.

\textbf{\textit{Tip 7: We suggest that in 4D hybrid parallelism, the order of process group dimensionality from low to high is TP, SP-Ulysses, SP-Ring, ZeRO-DP, PP.}}

\section{Experimental Results}

\subsection{Performance of Unified SP}
We evaluated the attention module's performance using SP-Unified on an L20 PCIe GPU cluster, benchmarking the throughput in TFLOPS. 
We replicated the parameter settings of the llama3-8B model. Table~\ref{tab:attn_l20} illustrates that on an 8xL20 setup, optimal throughput for 32K and 128K sequences was achieved with a ulysses-degree of 4 and ring-degree set to 2. 
This finding supports Tip 1: SP-Unified is well-suited for heterogeneous networks.
The lb-ring, an enhanced version of the standard Ring-Attention with load-balancing, outperformed the original. 
It consistently outperforms the basic ring attention implementation, with the advantage becoming more pronounced as the sequence length increases.

\begin{table}[h]
\centering
\small
\caption{Throughput (iters/sec) of SP-Unified on 8xL20 PCIe fwd-only (Ring-Degree $\times$ Ulysses-Degree=8)}
\label{tab:attn_l20}
\label{tab:performance}
\begin{tabular}{|c|c|c|c|c|c|c|c|}
\hline
group\_num & bs & seqlen & head\_num & head\_size & ulysses\_degree & basic-ring & lb-ring \\ 
\hline
4 & 1 & 8K & 32 & 128 & 8 & 57.346 & 57.098 \\
4 & 1 & 8K & 32 & 128 & 4 & \textbf{153.134} & 152.189 \\
4 & 1 & 8K & 32 & 128 & 2 & 415.5 & 454.93 \\
4 & 1 & 8K & 32 & 128 & 1 & 358.595 & 361.969 \\
\hline
4 & 1 & 32K & 32 & 128 & 8 & 15.229 & 14.262 \\
4 & 1 & 32K & 32 & 128 & 4 & 28.584 & \textbf{32.818} \\
4 & 1 & 32K & 32 & 128 & 2 & 44.348 & 62.754 \\
4 & 1 & 32K & 32 & 128 & 1 & 40.478 & 58.377 \\
\hline
4 & 1 & 128K & 32 & 128 & 8 & 2.563 & 2.586 \\
4 & 1 & 128K & 32 & 128 & 4 & 3.217 & \textbf{4.235} \\
4 & 1 & 128K & 32 & 128 & 2 & 3.399 & 5.476 \\
4 & 1 & 128K & 32 & 128 & 1 & 3.131 & 5.186 \\
\hline
\end{tabular}
\end{table}

As shown in Table~\ref{tab:attn_a100}, we repeated the attention benchmarking on an 8xA100-SXM4 NVLink node, the highest throughput for both 32K and 128K sequence lengths was achieved when the ulysses-degree is set to 8, that is the same as SP-Ulysses.
SP-Ulysses demonstrated a significant advantage over SP-Ring. 
This verified the argument in Sec.~\ref{sec:sp-unified} of SP-Ring, that although communication overhead can be hidden through overlapping with computation, it results in a reduction in computational efficiency.

\begin{table}[h]
\centering
\small
\caption{Throughput (iters/sec) of SP-Unified on 8xA100-SXM4 NVLink fwd-only (Ring-Degree $\times$ Ulysses-Degree=8)}
\label{tab:attn_a100}
\begin{tabular}{|c|c|c|c|c|c|c|c|}
\hline
group\_num & bs & seqlen & head\_num & head\_size & ulysses\_degree & basic-ring & lb-ring \\ \hline
4 & 1 & 32K & 32 & 128 & 8 & 135.569 & \textbf{136.375} \\
4 & 1 & 32K & 32 & 128 & 4 & 103.525 & 132.979 \\
4 & 1 & 32K & 32 & 128 & 2 & 91.365 & 132.979 \\
4 & 1 & 32K & 32 & 128 & 1 & 81.985 & 113.79 \\ 
\hline
4 & 1 & 128K & 32 & 128 & 8 & 2.782 & \textbf{2.785} \\
4 & 1 & 128K & 32 & 128 & 4 & 2.024 & 2.771 \\
4 & 1 & 128K & 32 & 128 & 2 & 1.73 & 2.89 \\
4 & 1 & 128K & 32 & 128 & 1 & 1.628 & 2.91 \\ 
\hline
\end{tabular}
\end{table}

\subsection{End-to-end SP Performance in Megatron-LM}
We have incorporated the SP-Unified method into Megatron-LM. Currently, Megatron-LM includes a preliminary version of SP-Ring, but lacks an implementation of SP-Ulysses. 
Our software is based on Megatron-LM commit 2196398, dated April 12, 2024. The SP-Ring is implemented using Megatron-LM's native Context Parallel, while SP-Ulysses is developed from our repository code. 
We conducted experiments using the docker image nvcr.io/nvidia/nemo:24.03. By default, we utilize ZeRO-1 for both Data Parallelism (DP) and Sequence Parallelism (SP), and consistently apply Sequence Parallelism Optimization for Tensor Parallelism (TP-sp). 
We do not employ gradient accumulation or activation recomputation. 
Our experimental setup includes two GPU nodes, each equipped with 8xA800 GPUs connected via 400GB/s NVLink, and node-to-node communication via 1.6 Tbps RDMA.
We have adjusted the MFU computation in Megatron-LM to account only for effective computations under causal masking. As a result, the MFU is notably lower when training with long sequences compared to the MFU figures printed by Megatron-LM.

\subsection{SP vs. DP}
Firstly, we compare the performance of SP and DP under the same LLAMA2-7B workload on a single node of A800 GPUs.
The global batch size is 8.
We use the SP-Unified, and pick the best ulysses-degree and ring-degree settings.
In a single node, ulysses-degree = 8 usually achieves the best performance.
As shown in Figure~\ref{fig:spvsdp}, DP outperforms SP-Unified across various input sequence lengths, which confirms our conclusion in Tip 2.

\begin{figure}[ht!]
    \centering
    \includegraphics[width=0.8\linewidth]{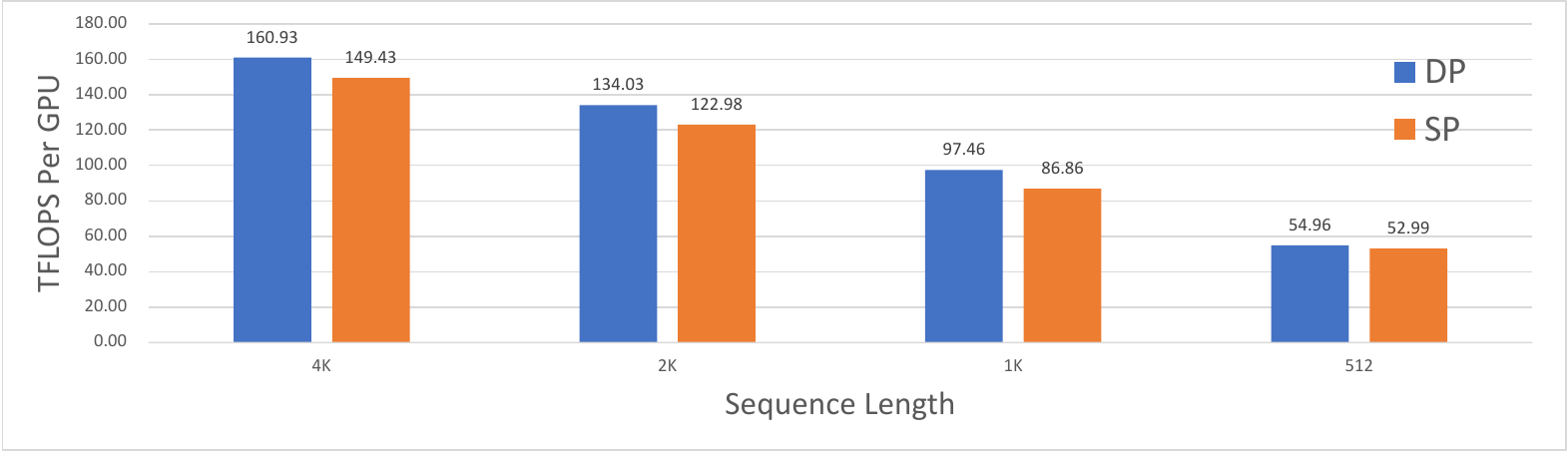}
    \caption{Comparison DP and SP on LLAMA2-7B Task with global bs=8.}
    \label{fig:spvsdp}
\end{figure}

\subsection{Hybrid SP and TP}

Table~\ref{tab:1node} presents the performance of the llama2-7B model on a single node of 8xA800 GPU 80GB. 
The longest sequence length that can be achieved is 64K, and the best performance is obtained when tp-degree=4 and ulysses-degree=2. 
It outperforms TP-sp-only by 10\%.
When seqlen=64K, SP-only will bring OOM issue, which echos our conclusion in Tip 5 that SP is less memory-efficient than TP-sp.

When global-bs=16 and seqlen=30K, SP-Ulysses delivers the optimal performance, significantly surpassing other SP and TP hybrid strategies. 
It is also 26\% better than TP-sp-only in throughput.
This indicates that, despite the similar communication cost of SP-Ulysses and TP-sp, 
SP-Ulysses is more communication efficient than TP-sp in practice in an NVLINK-connected setting, and communication patterns of hybrid SP-Ulysses and SP-Ring 
 sometimes bring better performance.

\begin{table}[ht!]
\small
\centering
\caption{LLAMA2-7B FLOPS per GPU of hybrid parallelism using TP and SP-Unified on a single 8xA800 NVLink node.}
\begin{tabular}{|c|c|c|c|c|c|c|}
\hline
seqlen & global-bs & tp-degree & ulysses-degree & ring-degree & FLOPS/GPU & MFU \\
\hline
64K & 1 & 4 & 2 & 1 & \textbf{154.49} & \textbf{0.50} \\
\hline
64K & 1 & 4 & 1 & 2 & 151.40 & 0.49 \\
\hline
64K & 1 & 8 & 1 & 1 & 141.85 & 0.45 \\
\hline
\hline
30K & 16 & 2 & 4 & 1 & 155.98 & 0.50 \\
\hline
30K & 16 & 2 & 1 & 4 & 147.77 & 0.47 \\
\hline
30K & 16 & 4 & 1 & 1 & 150.05 & 0.48 \\
\hline
30K & 16 & 1 & 8 & 1 & \textbf{163.42} & \textbf{0.52} \\
\hline
30K & 16 & 1 & 1 & 8 & 142.16 & 0.46 \\
\hline
30K & 16 & 8 & 1 & 1 & 129.12 & 0.41 \\
\hline
\end{tabular}
\label{tab:1node}
\end{table}

We employed a hybrid parallel strategy combining TP and SP-Unified to benchmark the training throughput of LLAMA3-8B across two nodes.
The results are shown in Table~\ref{tab:2node}.
In most of cases, the global batch size is fixed to be 1, aiming to maximize the sequence length. 
Since llama3-8B has only 8 heads, the maximum product of ulysses-degree and tp-degree is 8.
Table~\ref{tab:2node} presents throughput in FLOPS on 64K, 80K and 120K.

\begin{table}[ht!]
\centering
\small
\caption{LLAMA3-8B FLOPS per GPU of hybrid parallelism using TP and SP-Unified on two RDMA-connected 8xA800 NVLink nodes.}
\begin{tabular}{|c|c|c|c|c|c|c|}
\hline
seqlen & global-bs & tp-degree & ulysses-degree & ring-degree & FLOPS/GPU & MFU \\ \hline
64K & 1 & 1 & 8 & 2 & 136.31 & 0.44 \\
64K & 1 & 1 & 4 & 4 & \textbf{137.48} & \textbf{0.44} \\
64K & 1 & 1 & 2 & 8 & 129.44 & 0.41 \\
64K & 1 & 1 & 1 & 16 & 121.83 & 0.39 \\
\hline
64K & 1 & 8 & 1 & 2 & {129.75} & {0.42} \\
64K & 1 & 4 & 2 & 2 & 122.45 & 0.39 \\
64K & 1 & 2 & 4 & 2 & 87.67 & 0.28 \\
64K & 1 & 2 & 2 & 4 & 89.35 & 0.29 \\
64K & 1 & 4 & 1 & 4 & 122.57 & 0.39 \\
64K & 1 & 2 & 1 & 8 & 101.35 & 0.32 \\
\hline
64K & 2 & 8 & 1 & 1 & 141.20 & 0.45 \\ 
\hline
\hline
80K & 1 & 1 & 8 & 2 & 147.46 & 0.47 \\
80K & 1 & 1 & 4 & 4 & \textbf{148.90} & \textbf{0.48} \\
80K & 1 & 1 & 2 & 8 & 140.13 & 0.45 \\
80K & 1 & 1 & 1 & 16 & 132.86 & 0.43 \\
\hline
80K & 1 & 8 & 1 & 2 & 136.16 & 0.44 \\
80K & 1 & 4 & 2 & 2 & \textbf{137.49} & \textbf{0.44} \\
80K & 1 & 2 & 4 & 2 & 111.05 & 0.36 \\
80K & 1 & 2 & 2 & 4 & 110.81 & 0.36 \\
80K & 1 & 4 & 1 & 4 & 130.27 & 0.42 \\
80K & 1 & 2 & 1 & 8 & 121.14 & 0.39 \\
\hline
80K & 2 & 8 & 1 & 1 & 144.40 & 0.46 \\
\hline
\hline
120K & 1 & 4 & 2 & 2 & \textbf{152.51} & \textbf{0.49} \\
\hline
120K & 1 & 2 & 4 & 2 & 136.63 & 0.44 \\
\hline
120K & 1 & 8 & 1 & 2 & 145.92 & 0.47 \\
\hline
120K & 1 & 4 & 1 & 4 & 150.96 & 0.48 \\
\hline
\end{tabular}
\label{tab:2node}
\end{table}

The optimal performance for sequence lengths of 64K and 80K is achieved with SP-only without TP-sp, the optimal setting of both is ulysses-degree=4 and ring-degree=4. 
For sequence length is 64K and 80K, the unified SP outperforms the SP-Ring by 13\% and 12\%, respectively. 
These results echo our conclusion in Tip 1.

For sequence lengths are 64K and 80K, we can increase the global batch size (global-bs) for the TP+SP hybrid setting when tp-degree=8.
However, SP-only always has an OOM issue with global-bs=2.
Increasing the global-bs to 2,  TP+SP has a 2.7\% improvement in throughput to SP-only at the sequence length 64K.
But at sequence length 80K, TP+SP with global-bs as 2 is still worse than SP-only with global-bs as 1.

When the sequence length reaches 120K, the optimal performance is achieved with tp-degree=4 and ulysses-degree=2, reaching 152.51 TFLOPS, with an MFU of 0.49. 
At this sequence length, SP-only meets an OOM issue.
This confirms our Tip 5 again, which implies that the memory efficiency of SP is inferior to that of TP-sp.
It is noteworthy that two other TP+SP hybrid settings yield similar performance, 0.47 and 0.48 in MFU, which is quite close to the optimal.

\begin{table}[ht!]
\centering
\small
\caption{Exploring Upper Bound of Sequence Length for LLAMA3-8B using TP and SP-Unified on 2 8xA800 NVLink nodes.}
\begin{tabular}{|c|c|c|c|c|c|c|}
\hline
seqlen & global-bs & tp-degree & ulysses-degree & ring-degree & FLOPS/GPU & MFU \\ \hline
\rowcolor{gray!10}
160K & 1 & 4 & 2 & 2 & 158.64  & 0.51 \\ \hline
160K & 1 & 8 & 1 & 2 & 156.63 & 0.50  \\ \hline
\textbf{208K} & 1 & 8 & 1 & 2 & 147.26  & 0.47  \\ \hline
\rowcolor{gray!10}
160K & 1 & 4 & 1 & 4 & 159.37  & 0.51  \\ \hline
\rowcolor{gray!10}
190K & 1 & 4 & 1 & 4 & \textbf{157.08} & \textbf{0.50} \\ \hline
\end{tabular}
\label{tab:2node_longest}
\end{table}

We explored the upper bound of sequence length on 2 nodes, and the results are presented in Table~\ref{tab:2node_longest}. 
The largest sequence length was achieved with tp-degree=8 and ring-degree=2. 
At this point, SP-only could not run due to OOM.
Compared to SP, TP-sp is more memory-efficient, hence for training the longest sequences, parallel degrees are limited by 8 attention head numbers should all be assigned to TP-sp. 
It is noteworthy that we can achieve longer sequence lengths through activation optimization, such as activation recomputation~\cite{chen2016training}, and offloading~\cite{ren2021zero, fang2022parallel}, but these methods will harm MFU.

\subsection{Convergence}
We compared the convergence differences between USP and DP.
Using the same dataset, we tested the loss curves over 10K iterations on 4 GPUs.
We found that the curves for USP and DP completely overlapped, which validates that our modifications to RoPE and SP-Ring for load balancing are correct.

\begin{figure}[ht!]
    \centering
    \includegraphics[width=0.8\linewidth]{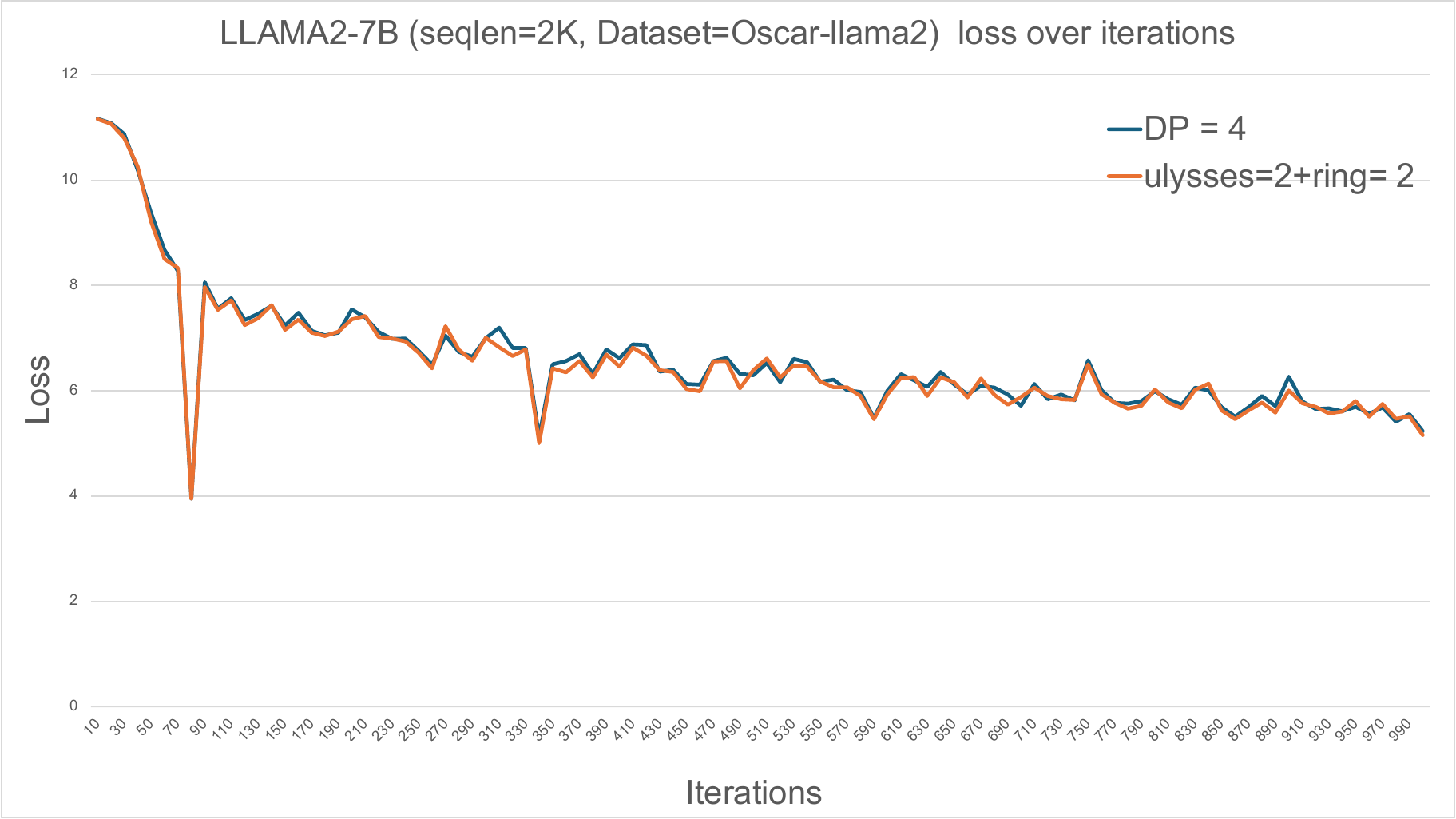}
    \caption{Comparison the loss of DP and USP on LLAMA2-7B with global bs as 4.}
    \label{fig:spvsdp}
\end{figure}

\section{Future Work}

\paragraph{SP on Large Scale Cluster}
We believe that SP is highly beneficial for extremely large scale LLM training tasks. 
Currently, for publicly disclosed large-scale model training tasks~\cite{jiang2024megascale, metallama3blog} over 10K GPUs, SP has not been utilized. 
This is because these training tasks were started before November 2023, when the SP methods were not yet mature.

Firstly, SP can introduce the dimension that can be partitioned from the sequence length, alleviating the constraints hindered by batch size limitations. 
MegaScale project~\cite{metallama3blog} uses a large global batch size to increase DP degree which impacts convergence. 
However, we can increase the SP degree instead of the DP degree to reduce the global batch size, thereby avoiding the convergence problem.

Secondly, increasing the SP degree can decrease the activation cost, allowing for longer model context length during training. 
Theoretically, the SP-Ring degree can be increased arbitrarily, whereas the TP degree is limited by the number of heads. 
Our experimental results on two nodes have not yet demonstrated this advantage of SP.

\paragraph{SP+ZeRO-3}
Megatron-LM does not officially support ZeRO-3, possibly because the TP-sp already reduces memory cost for parameters and gradients, which is also the target of ZeRO-3.
We have claimed that ZeRO-3 is highly compatible with SP in Tip 2.
Given that Megatron-LM already has an implementation of SP-Ring, ZeRO-3 becomes a necessary feature.

\paragraph{SP+MoE}
As more and more models transition to a Mixture-of-Experts (MoE) architecture, the research on combining SP with expert parallelism (EP) becomes increasingly significant.
The sequence parallelism employed on the attention module is decoupled with the FFN module.
Therefore, hybrid sequence parallelism can also be compatible with Mixture of Experts (MoE), as long as the All2All communication between the Attention and FFN modules is carefully designed.

\section{Conclusion}
In this paper, we propose an approach that unifies DeepSpeed-Ulysses and Ring-Attention for sequence parallelism. 
This method encompasses the capabilities of both techniques, broadening the applicability and delivering superior performance in some cases. 
We systematically analyzed the interplay between sequence parallelism and other established parallelism methods, deriving a set of best practice suggestions.
These suggestions have been validated through experimental results obtained from two GPU nodes.

\section{Acknowledgements}
We express our gratitude to Zilin Zhu from Tencent. 
The code utilized in our research was built from his GitHub repository, 
and he also owns the authorship of Figure~\ref{fig:lb} presented in this paper.

\bibliographystyle{unsrt}  
\bibliography{ref}  

\end{document}